%% file: main.tex
\documentclass[letterpaper, 10pt, conference]{ieeeconf}
\IEEEoverridecommandlockouts
\usepackage[pdftex]{graphicx}
 
\graphicspath{{../images/}}
\DeclareGraphicsExtensions{.pdf,.jpeg,.png}

\usepackage{amssymb}
\usepackage[natural]{xcolor}
\usepackage[cmex10]{amsmath}
\usepackage[tight,footnotesize]{subfigure}
\usepackage{url}
\usepackage{ulem} 
\usepackage{csquotes}
\usepackage[mode=buildnew]{standalone}
\usepackage{tikz}
\newcommand{\cmmnt}[1]{\ignorespaces}

\title{\LARGE \bf An Application-Driven Conceptualization of Corner Cases \\for Perception in Highly Automated Driving}

\author{Florian Heidecker$^{1}$,
        Jasmin Breitenstein$^{2}$,
        Kevin Rösch$^{3}$,
        Jonas Löhdefink$^{2}$, 
        Maarten Bieshaar$^{1}$,\\
        Christoph Stiller$^{4}$,
        Tim Fingscheidt$^{2}$,
        and
        Bernhard Sick$^{1}$
    \thanks{This paper is submitted to IEEE Intelligent Vehicles Symposium 2021.}
    \thanks{$^{1}$Florian Heidecker, Maarten Bieshaar and Bernhard Sick are with Intelligent Embedded Systems, University of Kassel, Wilhelmshöher Allee 73, 34121 Kassel, Germany, {\tt\small \{florian.heidecker, mbieshaar, bsick\}@uni-kassel.de}}
    \thanks{$^{2}$Jasmin Breitenstein, Jonas Löhdefink and Tim Fingscheidt are with Institute for Communications Technology, Technische Universität Braunschweig, Schleinitzstraße 22, 38106 Braunschweig, Germany, {\tt\small \{j.breitenstein, j.loehdefink, t.fingscheidt\}@tu-bs.de}}%
    \thanks{$^{3}$Kevin Rösch is with Mobile Perception Systems, FZI Research Center for Information Technology, Schönfeldstraße 8, 76131 Karlsruhe, Germany, {\tt\small kevin.roesch@fzi.de}}%
    \thanks{$^{4}$ Christoph Stiller is with Institute  of  Measurement  and  Control  Systems, Karlsruhe Institute of Technology, Engler-Bunte-Ring 21, 76131 Karlsruhe, Germany, {\tt\small stiller@kit.de}}%
}

\begin{document}

\maketitle

\newcommand{\fh}[1]{{\color{black} #1}}
\newcommand{\jb}[1]{{\color{black} #1}}
\newcommand{\jl}[1]{{\color{black} #1}}
\newcommand{\mb}[1]{{\color{black} #1}}
\newcommand{\kr}[1]{{\color{black} #1}}

\input{sections/ch0_abstract}

\IEEEpeerreviewmaketitle

\input{sections/ch1_introduction2}

\input{sections/ch2_distinction_between_cc}

\input{sections/ch4_toolchain}
\input{sections/ch5_sensors}
\input{sections/ch3_data_driven_corner_cases}
\input{sections/ch7_methods}
\input{sections/ch8_evaluation}
\input{sections/ch9_conclusion}

\input{sections/acknowledgment}

\bibliographystyle{IEEEtran}
\bibliography{IEEEabrv, bibliography}
\end{document}

%% file: sections/ch0_abstract.tex
\begin{abstract}
    \fh{Systems and functions that rely on machine learning (ML) are the basis of highly automated driving.} \jb{An essential task \fh{of such ML models is to reliably detect and interpret} unusual, new, and potentially dangerous situations. \mb{The detection of those situations, which we refer to as corner cases, is highly relevant for successfully \jl{developing}, applying, and validating automotive perception functions \fh{in future vehicles where multiple sensor modalities will be used.}}
    A complication for the development of corner case detectors is the lack of consistent definitions, terms, and corner case descriptions, especially when taking into account various automotive sensors.}
    \jb{In this work, we provide an application-\jl{driven} view of corner cases in highly automated driving. \fh{To achieve this goal, we first consider existing definitions from the general outlier, \fh{novelty,} anomaly, and out-of-distribution detection \fh{to show relations and differences to corner cases.} 
    Moreover, we extend an existing camera-\jl{focused} systematization of corner cases by adding RADAR (radio detection and ranging) and LiDAR (light detection and ranging) sensors. }
    For this, we describe an exemplary toolchain for data acquisition and processing, highlighting the interfaces of the corner case detection. We also \fh{define} a \jl{novel} level of corner cases, the \textit{method layer} corner cases, which appear due to uncertainty inherent in the methodology or the data distribution. }
\end{abstract}

%% file: sections/ch1_introduction2.tex
\section{Introduction}\label{sec:introduction}
    \jb{Perception is a challenging task in highly automated driving. \fh{Especially when, as in Figure \ref{fig:winter_cc}, low winter sun overexposes the camera, the LiDAR provides false information due to reflections from the partly slippery/icy road, and the RADAR has much noise in its data due to the snow.} Most importantly, to ensure safety, the environment perception methods need to operate reliably. This is crucial in driving situations deviating from what is considered ``normal'', and possibly even dangerous \fh{as in Fig.~\ref{fig:winter_cc}.} Such situations are generally termed as \textit{corner cases}, and their robust detection is necessary for reliable perception methods, and hence, safe highly automated driving. 
    \begin{figure}[htb]
        \centering
        \includegraphics[width=0.48\textwidth]{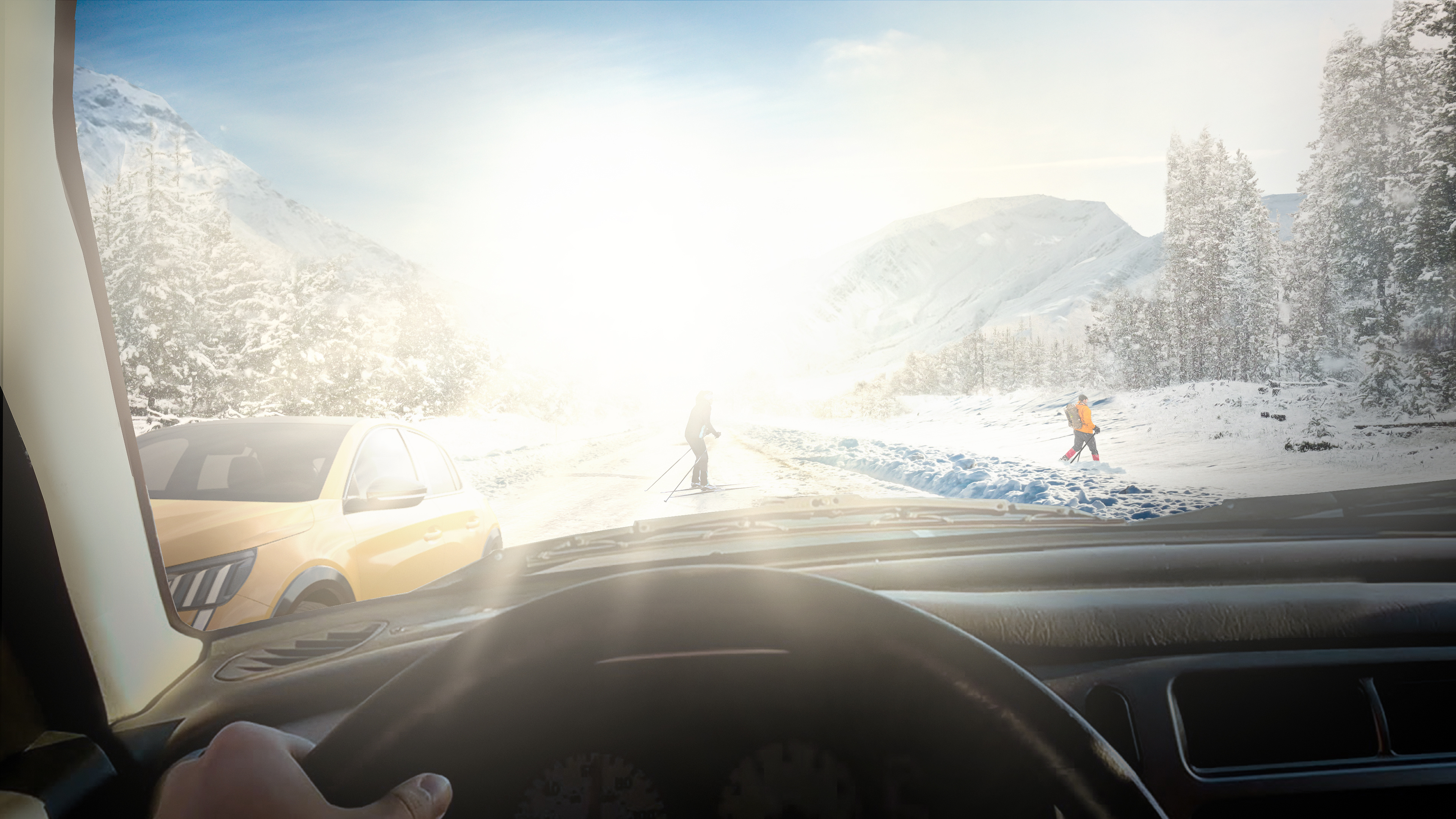}
        \vspace{0pt}
        \caption{Multiple corner cases: A winter scene with an icy,  slippery,  reflective  road,  combined  with  low  winter  sun and people on cross-country skis crossing the road.}
        \vspace*{-14pt}
        \label{fig:winter_cc}
    \end{figure}
    Next to the \jl{online} application, detection of corner cases is also required during development of machine learning (ML) perception methods. Here, they provide both the appropriate training, and the crucial test data to successfully develop and validate robust perception methods.\\
    While there exist many terms related to corner cases in literature, as well as many detection approaches, a general definition and description is missing. The lack of agreed technical terms and definitions makes detection of corner cases cumbersome. Previous work on corner case definitions and detection has mostly considered the camera sensor and visual perception. In this work, we consider corner cases of RADAR and LiDAR sensors in addition to the camera sensor.\\}
    \jb{The authors of \cite{Breitenstein2020} formulate a systematization of corner cases for visual perception in highly automated driving, where corner cases are categorized in levels. These levels are based on the type of situation they encompass and ordered by their theoretical detection complexity.
    We follow this approach and \kr{also consider corner cases on \textit{scene} \fh{(Fig.~\ref{fig:winter_cc})}, \textit{object} \fh{(e.g., people on cross-country skis)}, and \textit{domain level} \fh{(e.g., snowy winter)}, which we summarize into the \textit{content layer}. Additionally, for corner cases at \textit{scenario level}, we define the \textit{temporal layer}\fh{, e.g., the unusual movement of a person with cross-country skis compared to a pedestrian.} We grouped the corner cases, depending on whether they concern single image frames and point clouds (\textit{content layer}), or multiple consecutive ones (\textit{temporal layer}). }
    As we aim to provide a more comprehensive conceptualization of corner cases, including multi-modal sensor inputs, we distinguish on the lowest theoretical detection complexity between \text{pixel-,} and point-cloud-level corner cases, terming this the \textit{sensor layer}. We will provide a detailed description of the expanded systematization, and sensor-specific corner cases.\\}
    \fh{The main contributions of this work are:
    \begin{itemize}
        \item Conceptualization of a sensor-driven categorization of corner cases for camera, LiDAR, and RADAR into different \textit{layers} and \textit{levels}.
        \item Differentiation of \textit{single-} and \textit{multi-source} corner cases in consideration of different fusion levels.
        \item Introduction of \textit{method layer} corner cases generated by the method itself and depending on parameters such as topology, design, and deployment.
    \end{itemize}
    The remainder of this article is structured as follows:
    \jl{Section~\ref{sec:distinction_between_cc} gives an overview about recent work in the field of corner cases and their categorization. The processing toolchain in highly automated driving including camera, LiDAR, and RADAR sensor is presented in Section~\ref{sec:toolchain}. Section~\ref{sec:sensors} handles the individual sensor types in more detail and names specific corner cases. Section~\ref{sec:data-droven-corner-cases} particularly introduces the \textit{method layer} for corner cases, while Section~\ref{sec:methods_for_detection} shows various methods to detect corner cases. Existing datasets which can be used for corner case detection and potential evaluation methods are provided in Section~\ref{sec:evaluation}. Finally, Section~\ref{sec:conclusion} concludes the key message of the paper.}}

%% file: sections/ch2_distinction_between_cc.tex
\section{\jl{Distinction between Corner Cases and other Terms used in Literature}} \label{sec:distinction_between_cc}
    In literature, various terms are used to describe unexpected, or anomalous situations and events. In this section, we point out their distinctions and relations to corner cases in highly automated driving.
    
    \textbf{Edge Cases:}
    \fh{In the context of software and hardware testing, both terms \textit{corner case} and \textit{edge case} are used. 
    Edge cases are situations or parameters that occur rarely but are already taken into account during development \cite{Koopman2019}. This also applies to \textit{extreme cases} or \textit{boundary cases}, which are often not explicitly mentioned but are included in the edge case terminology.
    By identifying and handling edge cases, the situations or parameters lose the edge case status and are considered normal \cite{Koopman2019}. In contrast, a corner case results from the combination of several normal situations or parameters that coincide simultaneously, thus representing a rare or never considered case or scene \cite{Koopman2019}. Transferring this to highly automated driving, an example scene could be as shown in Fig.~\ref{fig:winter_cc}.

    In automated driving also entirely new situations can occur and are then considered as corner cases, not just combinations of already known ones.
    Similar to software testing, they can cease to exist once an appropriate number of examples of a particular corner case have been added to the training and validation data of a perception method.
    
    \textbf{ML viewpoint on corner cases:} In ML, we are interested in corner case situations to be able to validate, but more importantly, also to improve our system by re-training. The recorded data of such a situation can be used as part of the training data for the ML system, as the performance of a perception model on corner case data for tasks such as object detection, classification, and prediction is of great interest. A general ML system can perform poorly on such corner case data because the data contains novel situations, or effects, which are not (sufficiently) present in the system's training data. The identification of such corner cases is a major challenge, which also requires \kr{understanding the performance of} an ML model.}
    
\begin{figure}[t]
        \centering
        \includegraphics[width=0.2\textwidth]{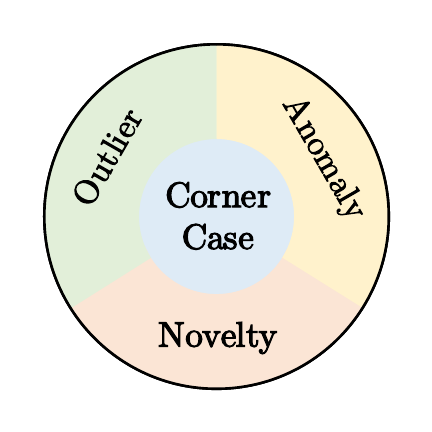} 
        \caption{
            The terms outlier, novelty, anomaly are often used in machine learning literature. Corner cases are strongly related to them and have an overlapping meaning with each of these terms.
        }
        \label{fig:cc_overlapping_meaning}
    \end{figure}

    \fh{In addition to the distinction between the terminology for corner cases in perception for highly automated driving and in software testing, the technical terms \textit{outliers}, \textit{anomalies} and \textit{novelties} from ML literature are related to the meaning of the term \textit{corner case}, \jl{see} Fig.~\ref{fig:cc_overlapping_meaning}.}
    \begin{figure*}[ht!]
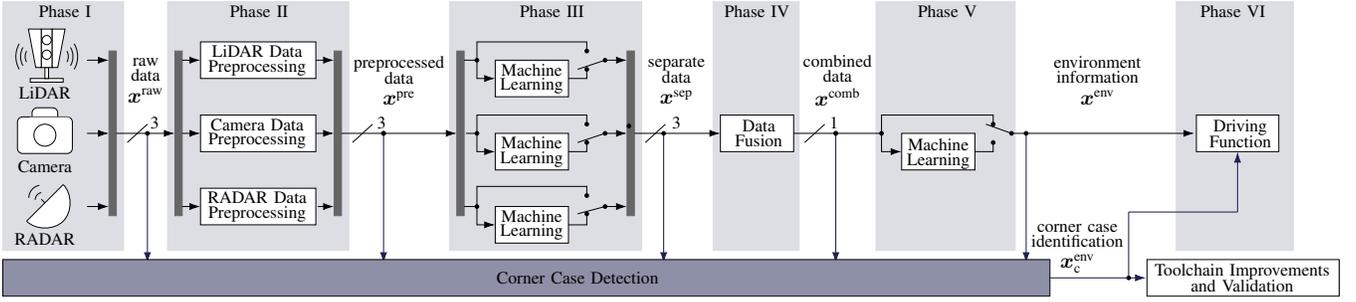

        \vspace{5pt}
        \centering
        \includestandalone[width=1.0\textwidth]{images_tex/pipeline}
        \vspace{0pt}
        \caption{
            \jl{Toolchain for data processing} showing the \jl{individual} sensors and processing steps with $I=3$ sensors: LiDAR, camera, and RADAR. \jl{The toolchain can be divided into the phases of data acquisition (I), data preprocessing (II), early machine learning (III), data fusion (IV), late machine learning (V), and application (VI).}
        }
        \label{fig:ml_toolchain}
    \end{figure*}
    
    \fh{\textbf{Outlier:} An outlier is defined as ``an observation which deviates so much from other observations as to arouse suspicions that it was generated by a different mechanism'' \cite{Hawkins1980}. In \cite{GRUHL2021}, an outlier is described as a legitimate observation of a known process that occurs in an area of low density and, e.g., represents an extreme value.
    We follow a similar understanding for corner cases, e.g., when the camera is maximally blinded or a LiDAR beam or RADAR \jb{impulse} is maximally reflected or absorbed by the environment.}
    
    \jb{\textbf{Anomalies:} For anomalies, multiple definitions can be found in literature (cf. \cite{Chandola2009,Pimentel2014,Breitenstein2020}). 
    While there are definitions focusing on anomalies as noisy data occurrences that prompt artifacts which in turn hinder data analysis \cite{Pimentel2014}, Jiang et al.\ \cite{Jiang2010}, e.g., first define normal events by a high frequency of occurrence, where consequently, anomalies are given as the opposite. 
    Other definitions consider anomalies as patterns that are not consistent with learned ones, or with general normal behavior \cite{Chandola2009,Popoola2012}. Moreover, there exists a categorization in contextual, collective, and point anomalies \cite{Chandola2009}.  
    For corner cases in automated driving, there is a deviation from normality that is manifested in non-conform behavior or patterns. The terms anomaly and corner case are almost used synonymously. Anomalies describe a deviation from normality. Hence, the term appears in the systematization of corner cases \cite{Breitenstein2020} as well. For corner cases, other factors also contribute to the terminology and definition, such as the relevance for the driving behavior \cite{Bolte2019b}. Moreover, corner cases can be more complex scenarios that become anomalous only in their entirety but do not consist of anomalous objects, similar again to the application of software testing.} 
    
    \fh{\textbf{Novelties:} Novelties appear as previously unseen instances or objects \cite{Chandola2009}. In \cite{GRUHL2021} the definition of novelties is somewhat broader, whereby novelties are described as a spatial or temporal agglomeration of anomalies or as a change in the distribution of an already known process. The appearance of new situations, objects, and movement patterns is an essential characteristic of corner cases, which does not simplify a clear distinction to novelties.}

%% file: images_tex/pipeline.tex
\begin{tikzpicture}[scale=1, every node/.style={scale=1.35}]

\usetikzlibrary{shapes.misc, positioning}
\usetikzlibrary{calc,shapes.geometric}
\usetikzlibrary{arrows}
\usetikzlibrary{arrows.meta}
\usetikzlibrary{patterns}
\definecolor{KIDTblue}{RGB}{28,33,77}
\definecolor{KIDTorange}{RGB}{245,124,20}
\definecolor{KIDTgreen}{RGB}{0,142,90}
\draw[fill=KIDTblue!50,KIDTblue!50]  (-5.55,3.3) rectangle (-5,-2.85);

\draw[fill=KIDTblue!15,KIDTblue!15]  (-7.6,3.75) rectangle (-4.3,-3.05);
\draw[fill=KIDTblue!15,KIDTblue!15]  (-3.1,3.75) rectangle (1.85,-3.05);
\draw[fill=KIDTblue!15,KIDTblue!15]  (4.6,3.75) rectangle (9.85,-3.05);
\draw[fill=KIDTblue!15,KIDTblue!15]  (11.8,3.75) rectangle (14.15,-3.05);
\draw[fill=KIDTblue!15,KIDTblue!15]  (16.25,3.75) rectangle (20.05,-3.05);

\draw[fill=KIDTblue!15,KIDTblue!15]  (24.45,3.75) rectangle (27.65,-3.05);
%
%
%
\draw[black,fill=white]  (-5.95,1.55) rectangle (-6.85,1.65);
\draw[black,fill=white]  (-6.35,1.95) rectangle (-6.45,1.65);
\draw[black,fill=white]  (-6.25,3) rectangle (-6.55,1.95);
\draw[fill=white]  (-6.4,2.7) ellipse (0.125 and 0.125);
\draw[fill=white]  (-6.4,2.3) ellipse (0.125 and 0.125);
\draw[fill=white] (-6.55,1.95) node[anchor=north]{}
  -- (-6.55,3) node[anchor=north]{}
  -- (-6.9,3) node[anchor=south]{}
   -- (-6.7,1.95) node[anchor=south]{}
  -- cycle;
\draw[fill=white] (-6.25,1.95) node[anchor=north]{}
  -- (-6.25,3) node[anchor=north]{}
  -- (-5.9,3) node[anchor=south]{}
  -- (-6.1,1.95) node[anchor=south]{}
  -- cycle;
\draw (-5.85,2.1) arc (-30:30:0.3);
\draw (-5.75,2) arc (-30:30:0.5);
\draw (-5.65,1.9) arc (-30:30:0.7);

\draw (-6.95,2.4) arc (150:210:0.3);
\draw (-7.05,2.5) arc (150:210:0.5);
\draw (-7.15,2.6) arc (150:210:0.7);

\draw[rounded corners,fill=white] (-6.8,0.85) rectangle (-6.1,0.4);
\draw[rounded corners,fill=white]  (-7.2,0.6) rectangle (-5.7,-0.4);
\draw[white,fill=white]  (-6.85,0.58) rectangle (-5.8,-0.2);
\draw  (-6.45,0.1) ellipse (0.3 and 0.3);

 \draw[fill=white] (-6.95,-2.25) .. controls (-6.7,-2.5) and (-6.2,-2.5) .. (-5.95,-2.25)
               .. controls (-5.7,-2) and (-5.7,-1.5) .. (-5.95,-1.25);
 \draw (-6.95,-2.25) -- (-5.95,-1.25);
 \draw (-6.45,-1.75) -- (-6.55,-1.65);
 \draw (-6.55,-1.47) arc (100:180:0.2);
  \draw (-6.6,-1.27) arc (100:180:0.35);


\draw[fill=black!60,black!60]  (-4.5,2.4) rectangle (-4.7,-2.1);
\draw[-{Latex[length=3mm, width=2mm]}] (-5.3,0.15) -- (-4.7,0.15);
\draw[-{Latex[length=3mm, width=2mm]}] (-5.3,-1.85) -- (-4.7,-1.85);
 \draw[-{Latex[length=3mm, width=2mm]}] (-5.3,2.15) -- (-4.7,2.15);
\draw[-{Latex[length=3mm, width=2mm]}] (-4.5,0.15) -- (-2.9,0.15);
\draw[fill=black!60,black!60]  (-2.7,2.4) rectangle (-2.9,-2.1);
\draw[-{Latex[length=3mm, width=2mm]}] (-2.7,0.15) -- (-2.2,0.15);
\draw[-{Latex[length=3mm, width=2mm]}] (-2.7,-1.85) -- (-2.2,-1.85);
\draw[-{Latex[length=3mm, width=2mm]}] (-2.7,2.15) -- (-2.2,2.15);

\draw[fill=white]  (-2.2,0.65) rectangle (0.95,-0.35);
\draw[fill=white]  (-2.2,2.65) rectangle (0.95,1.65);
\draw[fill=white]  (-2.2,-1.35) rectangle (0.95,-2.35);
\node[align=center,font=\linespread{0.7}\selectfont] at (-0.65,2.15) {LiDAR Data \\ Preprocessing};
\node[align=center,font=\linespread{0.7}\selectfont] at (-0.65,0.15) {Camera Data \\ Preprocessing};
\node[align=center,font=\linespread{0.7}\selectfont] at (-0.65,-1.85) {RADAR Data \\ Preprocessing};
\draw[fill=white]  (12,0.65) rectangle (14,-0.35);
\node[align=center,font=\linespread{0.7}\selectfont] at (13.05,0.15) {Data \\  Fusion};


\draw (-4.2,-0.1) -- (-3.65,0.4);
\node[align=center] at (-3.45,0.45) {3};

%

\draw[fill=white]  (25,0.65) rectangle (27.25,-0.35);
\node[align=center,font=\linespread{0.7}\selectfont] at (26.2,0.15) {Driving \\ Function};

\draw (1.95,-0.1) -- (2.5,0.4);
\node[align=center] at (2.75,0.45) {3};

\draw (9.95,-0.1) -- (10.5,0.4);
\node[align=center] at (10.8,0.45) {3};

\draw (14.3,-0.1) -- (14.85,0.4);
\node[align=center] at (15.1,0.45) {1};

\draw[fill=white]  (23.65,-3.3) rectangle (28.9,-4.3);
\node[align=center,font=\linespread{0.7}\selectfont] at (26.25,-3.8) {Toolchain Improvements  \\ and Validation};

\draw[fill=black!60,black!60]  (1.45,2.4) rectangle (1.65,-2.1);
\draw[-{Latex[length=3mm, width=2mm]}] (0.95,0.15) -- (1.45,0.15);
\draw[-{Latex[length=3mm, width=2mm]}] (1,-1.85) -- (1.45,-1.85);
\draw[-{Latex[length=3mm, width=2mm]}] (0.95,2.15) -- (1.45,2.15);
\draw[-{Latex[length=3mm, width=2mm]}] (1.65,0.15) -- (4.8,0.15);
\draw[fill=black!60,black!60]  (4.8,2.4) rectangle (5,-2.1);

 \begin{scope}[shift={(-2.4,2.65)}]
	\draw[fill=black]  (10.75,-0.25) ellipse (0.05 and 0.05) node [inner sep=0, outer sep=0] (lidar_switch_input1) {} {} {};
	\draw[fill=black]  (10.75,-0.75) ellipse (0.05 and 0.05) node [inner sep=0, outer sep=0] (lidar_switch_input2) {} {} {};
	\draw[fill=black]  (11.25,-0.5) ellipse (0.05 and 0.05) node [inner sep=0, outer sep=0] (lidar_switch_output) {} {} {};
	\draw (lidar_switch_output) -- (10.5,-0.75);
\end{scope}

\draw[fill=black] (5.35,2.15) ellipse (0.05 and 0.05) node [inner sep=0, outer sep=0] (lidar_shortcut_dot) {};
\draw (lidar_shortcut_dot) |- (6.85,2.65) -| (lidar_switch_input1);

 \begin{scope}[shift={(-2.4,0.65)}]
	\draw[fill=black]  (10.75,-0.25) ellipse (0.05 and 0.05) node [inner sep=0, outer sep=0] (camera_switch_input1) {} {} {};
	\draw[fill=black]  (10.75,-0.75) ellipse (0.05 and 0.05) node [inner sep=0, outer sep=0] (camera_switch_input2) {} {} {};
	\draw[fill=black]  (11.25,-0.5) ellipse (0.05 and 0.05) node [inner sep=0, outer sep=0] (camera_switch_output) {} {} {};
	\draw (camera_switch_output) -- (10.5,-0.75);
\end{scope}

\draw[fill=black] (5.35,0.15) ellipse (0.05 and 0.05) node [inner sep=0, outer sep=0] (camera_shortcut_dot) {};
\draw (camera_shortcut_dot) |- (6.85,0.65) -| (camera_switch_input1);

 \begin{scope}[shift={(-2.4,-1.3)}]
	\draw[fill=black]  (10.75,-0.25) ellipse (0.05 and 0.05) node [inner sep=0, outer sep=0] (radar_switch_input1) {} {} {};
	\draw[fill=black]  (10.75,-0.75) ellipse (0.05 and 0.05) node [inner sep=0, outer sep=0] (radar_switch_input2) {} {} {};
	\draw[fill=black]  (11.25,-0.55) ellipse (0.05 and 0.05) node [inner sep=0, outer sep=0] (radar_switch_output) {} {} {};
	\draw (radar_switch_output) -- (10.5,-0.75);
\end{scope}

\draw[fill=black] (5.35,-1.85) ellipse (0.05 and 0.05) node [inner sep=0, outer sep=0] (radar_shortcut_dot) {};
\draw (radar_shortcut_dot) |- (6.85,-1.3) -| (radar_switch_input1);

 \begin{scope}[shift={(-4.15,0.7)}]
	\draw[fill=white]  (10,-0.55) rectangle (12,-1.55);
	\node[align=center,font=\linespread{0.7}\selectfont] at (11,-1.05) {Machine\\Learning};
	\coordinate (camera_nn_in) at (10,-1.05) {} {} {} {} {} {} {} {} {} {} {} {};
	\coordinate (camera_nn_out) at (12,-1.05) {} {} {} {} {} {} {} {} {} {} {} {} {};
\end{scope}

 \begin{scope}[shift={(-4.15,-1.8)}]
	\draw[fill=white]  (10,-0.05) rectangle (12,-1.05);
	\node[align=center,font=\linespread{0.7}\selectfont] at (11,-0.55) {Machine\\Learning};
	\coordinate (radar_nn_in) at (10,-0.55) {} {} {} {} {} {} {} {} {} {} {} {} {};
	\coordinate (radar_nn_out) at (12,-0.55) {} {} {} {} {} {} {} {} {} {} {} {} {} {};
\end{scope}

 \begin{scope}[shift={(-4.15,2.6)}]
	\draw[fill=white]  (10,-0.45) rectangle (12,-1.45);
	\node[align=center,font=\linespread{0.7}\selectfont] at (11,-0.95) {Machine\\Learning};
	\coordinate (lidar_nn_in) at (10,-0.95) {} {} {} {} {} {} {} {} {} {} {} {};
	\coordinate (lidar_nn_out) at (12,-0.95) {} {} {} {} {} {} {} {} {} {} {} {} {};
\end{scope}

 \begin{scope}[shift={(6.95,0.7)}]
	\draw[fill=white]  (10,-0.55) rectangle (12,-1.55);
	\node[align=center,font=\linespread{0.7}\selectfont] at (11,-1.1) {Machine\\Learning};
	\coordinate (fused_nn_in) at (10,-1.05) {} {} {} {} {} {} {} {} {} {} {} {} {} {} {} {} {} {} {};
	\coordinate (fused_nn_out) at (12,-1.05) {} {} {} {} {} {} {} {} {} {} {} {} {} {} {} {} {} {} {} {};
\end{scope}

 \begin{scope}[shift={(8.7,0.65)}]
	\draw[fill=black]  (10.75,-0.25) ellipse (0.05 and 0.05) node [inner sep=0, outer sep=0] (fused_switch_input1) {} {} {};
	\draw[fill=black]  (10.75,-0.75) ellipse (0.05 and 0.05) node [inner sep=0, outer sep=0] (fused_switch_input2) {} {} {};
	\draw[fill=black]  (11.25,-0.5) ellipse (0.05 and 0.05) node [inner sep=0, outer sep=0] (fused_switch_output) {} {} {};
	\draw (fused_switch_output) -- (10.55,-0.25);
\end{scope}

\draw[fill=black] (16.4,0.15) ellipse (0.05 and 0.05) node [inner sep=0, outer sep=0] (fused_shortcut_dot) {};
\draw (fused_shortcut_dot) |- (17.95,0.65) -| (fused_switch_input1);

 \draw[-{Latex[length=3mm, width=2mm]}] (camera_switch_output) -- (9.45,0.15);
\draw[-{Latex[length=3mm, width=2mm]}] (5,-1.85) -| (radar_shortcut_dot) |- (radar_nn_in);
\draw[-{Latex[length=3mm, width=2mm]}] (radar_switch_output)  -- (9.45,-1.85);
\draw[-{Latex[length=3mm, width=2mm]}] (5,2.15) -| (lidar_shortcut_dot) |- (lidar_nn_in);
\draw (lidar_nn_out) -| (lidar_switch_input2);
\draw (radar_nn_out) -| (radar_switch_input2);
\draw (camera_nn_out) -| (camera_switch_input2);
\draw (fused_nn_out) -| (fused_switch_input2);
\draw[-{Latex[length=3mm, width=2mm]}] (lidar_switch_output)  -- (9.45,2.15);
\draw[-{Latex[length=3mm, width=2mm]}] (5.05,0.15) -| (camera_shortcut_dot) |- (camera_nn_in);

\draw[fill=black!60,black!60]  (9.45,2.4) rectangle (9.65,-2.1);
\draw[-{Latex[length=3mm, width=2mm]}] (9.65,0.15) -- (12,0.15);

  \draw[-{Latex[length=3mm, width=2mm]}] (14,0.15) -| (fused_shortcut_dot) |- (fused_nn_in);
 \draw[-{Latex[length=3mm, width=2mm]}] (fused_switch_output) -- (24.95,0.15);
 
 
\draw[fill=black] (15.15,0.15) ellipse (0.05 and 0.05) node [inner sep=0, outer sep=0] (fused_cc_dot) {};
 \draw[KIDTblue,thick,-{Latex[length=3mm, width=2mm]}] (fused_cc_dot) -- (15.15,-3.4);   
 \draw[fill=black] (15.15,0.15) ellipse (0.05 and 0.05) node [inner sep=0, outer sep=0] (fused_cc_dot) {};
  \draw[KIDTblue,thick,-{Latex[length=3mm, width=2mm]}]  (10.45,0.15) -- (10.45,-3.4);   
\draw[KIDTblue,thick,-{Latex[length=3mm, width=2mm]}]  (20.35,0.15) -- (20.35,-3.4); 
  \draw[KIDTblue,thick,-{Latex[length=3mm, width=2mm]}]  (-3.65,0.15) -- (-3.65,-3.4);   
  
 \draw[fill=black] (2.8,0.15) ellipse (0.05 and 0.05) node [inner sep=0, outer sep=0] (camera_cc_dot) {};
 \draw[KIDTblue,thick,-{Latex[length=3mm, width=2mm]}]  (camera_cc_dot) -- (2.8,-3.4);
  \draw[fill=black] (2.8,0.15) ellipse (0.05 and 0.05) node [inner sep=0, outer sep=0] (camera_cc_dot) {};
 
 \node[align=center] at (-6.45,1.25) {LiDAR}; 
\node[align=center] at (-6.45,-0.75) {Camera}; 
\node[align=center] at (-6.45,-2.75) {RADAR}; 
 \node[align=center,font=\linespread{0.7}\selectfont] at (10.8,1.6) {separate\\data \\[0.4em] \large $\boldsymbol{x}^\text{sep}$}; 
  \node[align=center,font=\linespread{0.7}\selectfont] at (15.2,1.6) {combined\\ data \\[0.4em] \large  $\boldsymbol{x}^\text{comb}$}; 
 
 \node[align=center,font=\linespread{0.7}\selectfont] at (-3.7,1.6) {raw \\ data \\[0.4em]  \large $\boldsymbol{x}^\text{raw}$};
  \node[align=center,font=\linespread{0.7}\selectfont] at (3.22,1.6) {preprocessed \\ data \\[0.4em]  \large $\boldsymbol{x}^\text{pre}$};

\node[align=center] at (-5.9,3.5) {Phase I}; 
\node[align=center] at (-0.6,3.5) {Phase II}; 
\node[align=center] at (7.4,3.5) {Phase III}; 
\node[align=center] at (13,3.5) {Phase IV}; 
\node[align=center] at (18.2,3.5) {Phase V}; 
\node[align=center] at (26,3.5) {Phase VI}; 
\draw[fill=KIDTblue!50]  (-7.6,-3.3) rectangle (21,-4.3);
\node[align=center] at (8.1,-3.8) {Corner Case Detection}; 

  \draw[KIDTblue,thick,-{Latex[length=3mm, width=2mm]}]  (21,-3.8) -- (23.625,-3.8);  
   \draw[KIDTblue,thick]  (23.15,-3.8) -- (23.15,-2);  
      \draw[KIDTblue,thick]  (23.15,-2) -- (26.125,-2);  
  
    \draw[KIDTblue,thick,-{Latex[length=3mm, width=2mm]}]  (26.125,-2) -- (26.125,-0.35); 

\node[align=center,font=\linespread{0.7}\selectfont] at (21.75,-3) {corner case \\ identification \\[0.4em] \large $\boldsymbol{x}^\text{env}_\text{c}$}; 
\node[align=center,font=\linespread{0.7}\selectfont] at (22.3,1.6) {environment \\ information \\[0.4em]  \large $\boldsymbol{x}^\text{env}$}; 
 
\draw[fill=black]  (20.35,0.15) ellipse (0.05 and 0.05);
\draw[fill=black]  (-3.65,0.15) ellipse (0.05 and 0.05);
\draw[fill=black]  (10.45,0.15) ellipse (0.05 and 0.05);
\draw[fill=black]  (9.5,0.35) ellipse (0.05 and 0.05);
\draw[fill=black]  (23.15,-3.8) ellipse (0.05 and 0.05);
\end{tikzpicture}

%% file: sections/ch4_toolchain.tex
    
    
 
\section{Multi-sensor Processing Toolchain}\label{sec:toolchain}
    \fh{Before dealing with the different types of corner cases, we detail the processing toolchain on which our considerations and arguments are based.} \jb{Fig.~\ref{fig:ml_toolchain} shows the different sensors of a prototypical automated vehicle in phase \textrm{I}. For the toolchain, we consider three sensors, namely LiDAR, camera, RADAR. After the acquisition, the raw data $\boldsymbol{x}^\text{raw} = \left(x_{i}^\text{raw}\right)$ for the different sensors enumerated by $i \in \mathcal{I} =\lbrace 1, 2, 3 \rbrace$ are first preprocessed in phase \textrm{II}. \fh{The connection between phase \textrm{I} and \textrm{II} and between all other phases and blocks are shown with an arrow. For clarity, if necessary, several connections are bundled into one, where the number above the arrow describes the number of connections.} As preprocessing depends heavily on the type of sensor, it is done separately for \fh{camera images and point clouds from LiDAR and RADAR.} Possible preprocessing operations are reshaping, aligning, \fh{or filtering.}\\
    \fh{Afterwards, depending on the type of data fusion, the data $\boldsymbol{x}^\text{pre} = \left(x_{i}^\text{pre}\right)$, $i \in \mathcal{I}$, first pass through ML algorithms separately in phase \textrm{III} or bypass them and are directly fused into the combined data $\boldsymbol{x}^\text{comb}$ in phase \textrm{IV}.} The latter is termed early data fusion. \fh{The so-called late data fusion is performed when the sensor data $x_{i}^\text{pre}$ are first processed separately in ML algorithms, and the outputs $\boldsymbol{x}^\text{sep} = \left(x_{i}^\text{sep}\right)$, $i \in \mathcal{I}$, are then fused afterward in phase \textrm{IV}.}
    Both processing options are visualized in Fig.~\ref{fig:ml_toolchain} by switches.
    In Section \ref{sec:sensors}, we regard early and late data fusion, and consider the possibilities of machine-learning-based fusion, where the phase \textrm{IV} and \textrm{V} from Fig.~\ref{fig:ml_toolchain} become indistinct and can even completely merge. The processing pipeline then outputs all environmental information $\boldsymbol{x}^\text{env}$ that are extracted in the pipeline.\\
    Furthermore, Fig.~\ref{fig:ml_toolchain} provides an interface description for corner case detection: Corner cases can appear at multiple points \fh{in the toolchain.} Therefore, if one aims to detect corner cases in the perception pipeline, the detection methods need to run in parallel to our depicted toolchain. Moreover, we require connections for the corner case detection from each processing phase \textrm{I} to \textrm{V}. \fh{This connection enables detecting corner cases in each step of the toolchain, providing all interim data from raw sensor values $\boldsymbol x_{i}^\text{raw}$, $i \in \mathcal{I}$, to fully extracted environment information $\boldsymbol{x}^\text{env}$.} Thus, we might detect\fh{, e.g.,} unknown objects after late fusion of camera and LiDAR, which deviate from each other, pixel errors directly in the raw data, or uncertainties from the ML module's output. \fh{The corner case detection has the second output of the toolchain and provides corner case information $\boldsymbol{x}^\text{env}_c$.} \jb{The corner case detection output is of the same size as the environment information $\boldsymbol{x}^\text{env}$ and provides for each entry a probability for a corner case appearing here. Both, corner case identification and environment information serve in phase \textrm{VI} (see Fig.~\ref{fig:ml_toolchain}) as input to the driving function, deciding on the appropriate driving behavior. In an offline (laboratory) application, phase \textrm{VI} is excluded, and instead, both $\boldsymbol{x}^\text{env}$ and $\boldsymbol{x}^\text{env}_c$ are used to reduce the data at hand intelligently. Note that the formulation of this toolchain can be arbitrarily extended to include more than three sensors by adding them in parallel to the existing ones.}}
 

%% file: sections/ch5_sensors.tex
\section{Sensor-Driven View of Corner Cases}\label{sec:sensors}
    \begin{table*}[t]
    \vspace{5pt}
        \centering
        \includestandalone[width=1.08\textwidth]{images_tex/table_coca}
        \caption{Categorization of camera-, LiDAR-, and RADAR-based \textit{single-source} corner cases. Example situations are given for the individual corner cases on the \textit{sensor layer}, \textit{content layer}, and \textit{temporal layer}. \textit{Method layer} corner cases are not shown in this table.}
        \label{tab:cc_camera_lidar_radar_fusion}
    \end{table*}

    \fh{The available data or the used algorithms do not only influence the occurrence of corner cases in ML. The application, in this case highly automated driving, the sensor technology, and the installation position have a significant \kr{impact}. Before discussing corner cases for different types of sensors, we briefly review a \kr{hypothetical} research vehicle as of today that could be the basis for highly automated vehicles in the future. It is equipped with a stereo camera, LiDAR sensors on the roof, and RADAR sensors on the front and rear. The field of view (FOV) of the different sensors overlaps mostly so that the environment is covered in most parts in each direction with at least two types of sensors. Based on this setup, there are sensor-specific corner cases for each sensor type, which is discussed in more detail below, where we extend a previous systematization \cite{Breitenstein2020} of corner cases for visual perception to the LiDAR and RADAR sensor.\\}
    \jb{The systematization \cite{Breitenstein2020} classifies corner cases into different levels, ordered by their theoretical complexity of detection. The highest detection complexity is given for \textit{scenario level} corner cases. These patterns can be observed throughout an image sequence and further subdivided into anomalous, novel, and risky scenarios based on their potential for collision and their observability during training. \textit{Scene level} corner cases are observed on single images and describe known objects in either unseen quantities or locations. On the \textit{object level}, unknown objects are observed in single images, while on the \textit{domain level}, corner cases arise due to the inability of the world model to explain its observations, i.e., a domain shift. The pixel level is the lowest in \cite{Breitenstein2020} and includes corner cases resulting from local and global outliers in the camera hardware.\\}
    \fh{In this work, we now extend and modify the vision-oriented systematization \cite{Breitenstein2020} by including other sensors and introducing the additional notion of a layer, see Table~\ref{tab:cc_camera_lidar_radar_fusion}.
    For clarity, the \textit{sensor layer}, \textit{content layer}, and \textit{temporal layer} are introduced at the top level. The \textit{temporal layer} includes corner cases with a temporal context, thereby corresponding to the \textit{scenario level} in \cite{Breitenstein2020}. On the other hand, the \textit{content layer} comprises the \textit{domain}, \textit{object}, and \textit{scene level} from \cite{Breitenstein2020} and thus contains corner cases that result (a) from the data at a specific point in time. The four levels \textit{scenario}, \textit{scene}, \textit{object}, and \textit{domain} are existing in signals from LiDAR and RADAR sensors as they exist in camera signals, as such corner cases appear in point clouds and sequences of point clouds in a similar way. However, it is important to consider that corner cases of a certain level can only be transferred from one sensor to another to a limited extent, if at all. A corner case can therefore exist for one sensor but not for a different sensor. Finally, the \textit{sensor layer} describes corner-cases that can be traced back to hardware errors or physical properties. This results in the newly introduced \textit{hardware level} and \textit{physical level}.\\
    Corner cases resulting from a single sensor are called \textit{single-source} corner cases. We have already discussed the categorization of these corner cases into layers and levels.
    However, the fusion of sensor data can cause \textit{multi-source} corner cases, which we consider as a separate category. Since they can appear more or less in any of the levels, we do not highlight them in Table~\ref{tab:cc_camera_lidar_radar_fusion}.}
    
    \subsection{Camera}\label{subsec:camera}
    \kr{In the context of highly automated driving, vehicles are often equipped with many different camera systems.} \fh{Mono and fisheye cameras are often used to cover the areas to the left, to the right, and behind the vehicle. Stereo, or rarely trifocal cameras, are used mostly to cover the area in front of the vehicle \cite{Winner.2015}. Regardless of the camera system, pixel errors such as dead pixel, a dirty camera lens, or overexposure can lead to a corner case at \textit{hardware level}, or \textit{physical level}, see Table~\ref{tab:cc_camera_lidar_radar_fusion}. Unique camera properties or functions, such as calculating the depth image of a stereo or trifocal camera, can cause corner cases at \textit{hardware level}. This group of corner cases can be summarized with corner cases at the \textit{sensor layer}.\\
    However, many corner cases are found in the image data itself and are often system-independent. Image data have a high information density. This fact is also reflected in various methods, such as object recognition and classification, contour estimation, recognition of the direction of gaze of the person, gesture recognition, traffic sign recognition, extraction of weather conditions and many more.
    On the other hand, this abundance of information and variability of image content also results in the most diverse corner cases. These corner cases can be divided into \textit{domain level}, e.g., different traffic signs in Germany and the U.S.A., \textit{object level}, such as animals and never-seen-before objects, and \textit{scene level}, where it can be a new situation such as a tree lying on the road. If the depth image is also available, the spectrum is extended by the possibility to determine the position of an object, which results in a further type of \textit{scene level} corner case. However, all these cases belong to the \textit{content layer}.\\
    If a camera captures a scene in a video sequence, it consists of several individual images 
    and the corner case can be represented by either a single frame, or by several consecutive frames. A corner case in a particular frame may require a temporal component as, e.g., in the case of a pedestrian's movement prediction, and is therefore a temporal-based corner case categorized on the \textit{scenario level}, situated at the \textit{temporal layer}.\\}

    \subsection{LiDAR}\label{subsec:lidar}
    \fh{There are different LiDAR systems in practice. Based on the technique, LiDARs are divided into two groups: the LiDAR system consisting of moving parts and the solid-state LiDAR system. In any vehicle, the sensors are exposed to vibrations, which leads to an increased risk of damage for LiDAR sensors with moving parts. In case of such or other hardware defects independent of the used LiDAR technology, the resulting corner cases can be summarized as \textit{hardware level} corner cases, see Table~\ref{tab:cc_camera_lidar_radar_fusion}. In addition to the measuring points, which result in a point cloud, the LiDAR also provides an intensity value for each measuring point. The intensity of the reflected beam can cause corner cases on the \textit{physical level}. For example, the intensity value can be very low for a dark-painted object or extremely high in the case of a mirror or license plate, leading to a wrong interpretation.\\
    The simplest type of LiDAR is the line laser, which can only measure one plane and depends heavily on the alignment. 
    If the LiDAR has several planes, the FOV expands vertically.
    \jb{A 360$^\circ$ LiDAR allows to cover all directions and} makes it possible to determine the width, height, and distance to each visible object around the vehicle.
    In all three LiDAR variants, corner cases occur on the \textit{content layer}. The corner cases can be quite different and corner cases in a LiDAR with several planes do not necessarily have to occur in a 360$^\circ$ LiDAR. On the \textit{domain level}, corner cases can arise, for example, because the road markings that are largely recognizable with a LiDAR in Germany cannot be perceived by a LiDAR in a different country because the markings are made of a different material. Dust or smoke clouds (extreme examples excepted) are a wonderful example for corner cases on \textit{object level}, since in comparison to camera and radar the laser beam of the LiDAR can be reflected by the particles. Corner cases can also arise at the scene level. For example, objects already known to a system that are in an unusual position, e.g., stacked scrap cars in a junkyard or a sweeper on the sidewalk, can cause corner cases.\\
    When a time sequence of measurements or point clouds is processed, corner cases at the \textit{temporal layer} arises. For example, this is expressed in the form of new movement patterns of people (e.g., movement of the extremities) or vehicles (e.g., the inclination of a moving motorcycle). We refer to this \kr{phenomenon} as \textit{scenario level} corner cases.}
    
    \subsection{RADAR}\label{subsec:radar}
    \fh{RADAR sensors for highly automated vehicles can be divided into short, mid, and long-range. The FOV becomes narrower with increasing range. For this reason, short and long-range sensors are combined often into one physical sensor system \cite{Winner.2015}. A RADAR has no moving parts as some LiDAR have, which can be damaged by vibration, or a lens like a camera that can be scratched. Nevertheless, this does not protect against a hardware defect, which can lead to a \textit{hardware level} corner case at the \textit{sensor layer}, see Table~\ref{tab:cc_camera_lidar_radar_fusion}. Corner cases resulting from interferences, e.g., from RADAR sensors of other vehicles, are summarized under the term \textit{physical level} corner cases. Further corner cases of the \textit{physical level} can be caused by a too small reflective surface of the object, making detection much worse or impossible.\\
    \textit{Object level} RADAR corner cases appear in acquired RADAR data and are not caused by poor accuracy or too small reflection surfaces. Multiple reflections can cause, for example, the so-called ghost detection of objects \cite{Winner.2015}. An example of this is if one side of a truck's tailgate is open and the other one is closed, the RADAR impulse is reflected several times in the trailer until the RADAR sensor receives the impulse again. A major advantage of RADAR sensors is the excellent determination of the relative speed of dynamic objects. On the other hand, different weather conditions affect the data and can lead to corner cases at the \textit{domain level}. For example, corner cases could arise at \textit{scene level} because the RADAR does not sufficiently detect large branches with leaves or trees that have fallen on the road. All these various RADAR corner cases are summarized as \textit{content layer} corner cases.\\
    Corner cases that result from a sequence of RADAR point clouds are summarized under the term \textit{scenario level} corner cases and belong to the \textit{temporal layer}, as previously for camera and LiDAR. As with the other sensors, corner cases are primarily addressed resulting from the trajectory of objects, e.g., movement patterns.}
    
    \subsection{Sensor Fusion}\label{subsec:fusion}
    While highly automated vehicles will be equipped with various sensors, each sensor \kr{has its} advantages and limitations. Henceforth, the data fusion module of the perception processing toolchain in Fig. \ref{fig:ml_toolchain} is crucial for the environment perception. Fig. \ref{fig:ml_toolchain} highlights two possibilities for data fusion in phase \textrm{IV}: early and late fusion. In early data fusion, the preprocessed data are directly fused in phase \textrm{IV}, bypassing phase \textrm{III}, i.e., $\boldsymbol{x}^\text{sep}=\boldsymbol{x}^\text{pre}$.
    Therefore, a network can learn to extract all information present in the raw data \cite{Feng2020}. 
    
    In late data fusion, information from the preprocessed data $\boldsymbol{x}^\text{pre}$ are extracted separately in the ML methods in phase \textrm{III}, and their output $\boldsymbol{x}^\text{sep}$ is combined \cite{Feng2020}.
    
    The so-called cross fusion \cite{Caltagirone2019}, or middle fusion \cite{Feng2020}, fuse the data, e.g., in intermediate layers of neural networks, hence blurring the distinct borders between the phases in Fig.~\ref{fig:ml_toolchain}, resulting in higher flexibility. ML approaches to data fusion can then also take the raw data as input and directly output environment information \cite{Qi2018}. However, there are many options on how and where to fuse the data \cite{Feng2020}.\\
    If the data from two or more sensors are merged, each sensor provides data that may well overlap in its features. If such an overlap leads to ambiguities that cannot be resolved, we summarize such situations under the term \textit{multi-source} corner cases. A \textit{multi-source} corner case occurs, for example, with a dust cloud: The LiDAR detects the dust cloud on the object level, while the camera and RADAR hardly perceive it or even not at all. 
    
    There is the potential that \textit{multi-source} corner cases in the data  fusion serve as a corner case detector themselves. For example, during sunset, visual perception of the camera sensor deteriorates. However, both other sensors, RADAR and LiDAR are not influenced by this corner case for visual perception. If a person crosses the street in front of the ego-vehicle in this situation, ambiguities appear in the data fusion. This leads to the critical question of trustworthiness of the perception methods.

%% file: images_tex/table_coca.tex
\begin{tikzpicture}[scale=1, every node/.style={scale=1.32}]
\usetikzlibrary{decorations.pathreplacing}
\definecolor{sensor}{RGB}{8,134,46}
\definecolor{content}{RGB}{10,171,231}
\definecolor{temporal}{RGB}{170,48,34}

\draw[sensor!20,fill=sensor!20]  (-3,-3) rectangle (8,6.5);
\draw[content!20,fill=content!20]  (8,-3) rectangle (24.5,6.5);
\draw[content!20,fill=temporal!20]  (24.5,-3) rectangle (30,6.5);

\draw  (19,-0.5) -- (24.5,-0.5) {};
\draw  (19,2)  {} -- (24.5,2) {};
\draw  (13.5,-0.5) -- (19,-0.5) {};
\draw  (13.5,2) -- (19,2) {};
\draw  (13.5,-0.5) -- (8,-0.5) {};
\draw  (13.5,2) -- (8,2) {};
\draw  (-3,-0.5) -- (8,-0.5) {};
\draw  (-3,2) -- (8,2) {};
\draw  (-6.15,4.5) -- (-3,4.5)  {};
\draw  (-6.15,2) -- (-3,2)  {};
\draw  (-6.15,-0.5) -- (8,-0.5)  {};
\draw  (-6.15,-3) -- (8,-3)  {};


\draw[black,fill=white]  (-4.2,2.9) rectangle (-5.1,3);
\draw[black,fill=white]  (-4.6,3.3) rectangle (-4.7,3);
\draw[black,fill=white]  (-4.5,4.3) rectangle (-4.8,3.3);
\draw[fill=white]  (-4.65,4) ellipse (0.125 and 0.125);
\draw[fill=white]  (-4.65,3.65) ellipse (0.125 and 0.125);
\draw[fill=white] (-4.8,3.3) node[anchor=north]{}
  -- (-4.8,4.3) node[anchor=north]{}
  -- (-5.15,4.3) node[anchor=south]{}
   -- (-4.95,3.3) node[anchor=south]{}
  -- cycle;
\draw[fill=white] (-4.5,3.3) node[anchor=north]{}
  -- (-4.5,4.3) node[anchor=north]{}
  -- (-4.15,4.3) node[anchor=south]{}
  -- (-4.35,3.3) node[anchor=south]{}
  -- cycle;
\draw (-4.1,3.45) arc (-30:30:0.3);
\draw (-4,3.35) arc (-30:30:0.5);
\draw (-3.9,3.25) arc (-30:30:0.7);

\draw (-5.2,3.75) arc (150:210:0.3);
\draw (-5.3,3.85) arc (150:210:0.5);
\draw (-5.4,3.95) arc (150:210:0.7);

\draw[rounded corners,fill=white] (-5,1.75) rectangle (-4.3,1.3);
\draw[rounded corners,fill=white]  (-5.4,1.5) rectangle (-3.9,0.5);
\draw[white,fill=white]  (-5.05,1.48) rectangle (-4,0.7);
\draw  (-4.65,1) ellipse (0.3 and 0.3);

 \draw[fill=white] (-5.2,-1.8) .. controls (-4.95,-2.05) and (-4.45,-2.05) .. (-4.2,-1.8)
               .. controls (-3.95,-1.55) and (-3.95,-1.05) .. (-4.2,-0.8);
 \draw (-5.2,-1.8) -- (-4.2,-0.8);
 \draw (-4.7,-1.3) -- (-4.8,-1.2);
 \draw (-4.8,-1.02) arc (100:180:0.2);
  \draw (-4.85,-0.82) arc (100:180:0.35);
  
  \node at (-0.5,5) { \textbf{Hardware Level}};
\node at (-0.25,4.1) {Laser Error};
\node[align=center, text width=7cm,font=\linespread{0.8}\selectfont] at (1.05,2.95) { \textit{\begin{itemize} \item Broken mirror\\ \vspace*{-2.5pt} \item Misaligned actuator \\  \end{itemize}}};
\node at (-0.25,1.6) {Pixel Error};
\node[align=center, text width=7cm,font=\linespread{0.8}\selectfont] at (1.1,0.55) { \textit{\begin{itemize} \item Dead pixel \\ \vspace*{-2.5pt}  \item Broken lense \\ \end{itemize}}};

\node at (-0.25,-0.95) {Impulse Error};

\node[align=center, text width=7cm,font=\linespread{0.8}\selectfont] at (1.1,-2.1) { \textit{\begin{itemize} \item Low voltage \\ \vspace*{-2.5pt} \item Low temperature \\  \end{itemize}}};

  \node at (5,5) { \textbf{Physical Level}};
\node[align=center,font=\linespread{0.8}\selectfont] at (5.25,3.95) {Beam-Based \\ Corner Case};
\node[align=center, text width=7cm,font=\linespread{0.8}\selectfont] at (6.6,0.55) { \textit{\begin{itemize} \item Dirt on lense \\ \vspace*{-2.5pt} \item Overexposure \\ \vspace*{-2.5pt}  \end{itemize}}};

\node[align=center,font=\linespread{0.8}\selectfont] at (5.25,1.45) {Pixel-Based \\  Corner Case};
\node[align=center, text width=7cm,font=\linespread{0.8}\selectfont] at (6.6,2.95) { \textit{\begin{itemize} \item \textit{Black cars} \textit{disappear} \\ \vspace*{-2.5pt} \item \ldots \end{itemize}}};
\node[align=center,font=\linespread{0.8}\selectfont] at (5.25,-1.05) {Impulse-Based \\  Corner Case};
\node[align=center, text width=7cm,font=\linespread{0.8}\selectfont] at (6.6,-2.1) { \textit{\begin{itemize} \item \textit{Interference} \\ \vspace*{-2.5pt} \item \ldots \end{itemize}}};

  
  \node [align=left,font=\linespread{0.8}\selectfont] (domain) at (10.5,5) { \textbf{Domain Level}};
\node[align=center,font=\linespread{0.8}\selectfont] at (10.75,3.95) {Domain Shift \\ on Single Point Cloud};
\node[align=center, text width=7cm,font=\linespread{0.8}\selectfont] at (12.1,0.45) { \textit{\begin{itemize} \item Location (EU-U.S.A.) \\ \vspace*{-2.5pt} \item \ldots \end{itemize}}};

\node[align=center,font=\linespread{0.8}\selectfont] at (10.75,1.45) {Domain Shift \\ on Single Frame};
\node[align=center, text width=7cm,font=\linespread{0.8}\selectfont] at (12.1,3.15) { \textit{\begin{itemize} \item Shape of Road \\ markings \\ \vspace*{-2.5pt} \end{itemize}}};

\node[align=center,font=\linespread{0.8}\selectfont] at (10.75,-1.05) {Domain Shift \\ on Single Point Cloud};
\node[align=center, text width=7cm,font=\linespread{0.8}\selectfont] at (12.15,-1.95) { \textit{\begin{itemize} \item Weather, e.g., \\ snow, rain, etc. \\  \end{itemize}}};

  \node at (16,4.95) { \textbf{Object Level}};
\node[align=center,font=\linespread{0.8}\selectfont] at (16.3,3.95) {Single-Point Anomaly \\  on Single  Point Cloud};
\node[align=center, text width=7cm,font=\linespread{0.8}\selectfont] at (17.6,2.95) { \textit{\begin{itemize} \item Dust cloud \\ \vspace*{-2.5pt} \item \ldots \end{itemize}}};

\node[align=center,font=\linespread{0.8}\selectfont] at (16.25,1.45) {Single-Point Anomaly \\ on Single Frame};
\node[align=center, text width=7cm,font=\linespread{0.8}\selectfont] at (17.6,0.45) { \textit{\begin{itemize}  \item Animal \\ \vspace*{-2.5pt} \item \ldots  \end{itemize}}};

\node[align=center,font=\linespread{0.8}\selectfont] at (16.25,-1.05) {Single-Point Anomaly \\  on Single  Point Cloud};
\node[align=center, text width=7cm,font=\linespread{0.8}\selectfont] at (17.6,-2.1) { \textit{\begin{itemize}  \item Lost objects \\ \vspace*{-2.5pt} \item \ldots \end{itemize}}};

  \node at (21.5,5) { \textbf{Scene Level}};
\node[align=center,font=\linespread{0.8}\selectfont] at (21.75,3.85) {Contextual/Collective \\ Anomaly on Single \\  Point Cloud };
\node[align=center, text width=7cm,font=\linespread{0.8}\selectfont] at (23.15,3.15) { \textit{\begin{itemize} \item Sweeper cleaning\\ the sidewalk \\ \vspace*{-2.5pt} \end{itemize}}};

\node[align=center,font=\linespread{0.8}\selectfont] at (21.7,1.3) {Contextual/Collective \\ Anomaly on \\ Single Frame};
\node[align=center, text width=7cm,font=\linespread{0.8}\selectfont] at (23.1,0.45) { \textit{\begin{itemize} \item People on a billboard \\ \vspace*{-2.5pt} \item \ldots \end{itemize}}};

\node[align=center,font=\linespread{0.8}\selectfont] at (21.75,-1.15) {Contextual/Collective \\ Anomaly on Single \\  Point Cloud };
\node[align=center, text width=7cm,font=\linespread{0.8}\selectfont] at (23.1,-2.1) { \textit{\begin{itemize} \item Demonstration \\ \vspace*{-2.5pt} \item Tree on street \end{itemize}}};

  \node at (15.95,6) {\textbf{Content Layer}};
    \node at (2.5,6) {\textbf{Sensor Layer}};
      \node at (27.1,5.95) {\textbf{Temporal Layer}};

  \node[align=center,font=\linespread{0.8}\selectfont] at (-4.55,-0.05) {Camera-based\\ corner cases};
    \node[align=center,font=\linespread{0.8}\selectfont] at (-4.55,-2.55) {RADAR-based \\ corner cases};
      \node[align=center,font=\linespread{0.8}\selectfont] at (-4.55,2.45) {LiDAR-based \\ corner cases};

\draw[black,very thick]  (8,-3) edge (8,6.5);
\draw[black,very thick]  (8,-3) edge (24.5,-3);
\draw[black,very thick]  (24.5,6.5) edge (24.5,-3);
\draw[black,very thick]  (-3,6.5) edge (30,6.5);
\draw[black,very thick]  (8,4.5) edge  (24.5,4.5);
\draw[black,very thick]  (8,5.5) edge (24.5,5.5);
\draw[black,very thick]  (19,5.5) edge (19,-3) ;
\draw[black,very thick]  (13.5,5.5) edge (13.5,-3) ;
\draw[black,very thick]  (30,-3) edge (24.5,-3);
\draw[black,very thick]  (24.5,6.5) edge (24.5,-3);
\draw[black,very thick] (30,6.5) edge (30,-3);
\node (v20) at (-3,6.5) {};
\draw[black,very thick]  (-3,5.5) edge (8,5.5);
\draw[black,very thick]   (-3,6.5) edge (-3,-3);
\draw[black,very thick]  (-6.15,-3) edge (8,-3);
\draw[black,very thick]  (2.5,-3) edge  (2.5,5.5);
\draw[black,very thick]  (-6.15,4.5)  edge (8,4.5);
\draw[black,very thick]   (-6.15,6.5) edge (-6.15,-3);
\draw[black,very thick]   (-6.15,6.5) edge (-3,6.5);

\draw[black,thick]  (24.5,4.5) edge (30,4.5) ; 
\draw[black,thick]  (24.5,5.5) edge (30,5.5) ;

  \node at (27,5) { \textbf{Scenario Level}};
\node[align=center,font=\linespread{0.8}\selectfont] at (27.3,3.45) {Corner Cases \\ on Multiple \\ Point Clouds \\ and Frames };


\node[align=center, text width=7cm,font=\linespread{0.8}\selectfont] at (29.25,0.9) { \textit{\begin{itemize} \item Person breaks \\ traffic rule \\ \vspace*{-2.5pt}  \item Overtaking a \\ cyclist \\ \vspace*{-2.5pt}  \item Car accident \\ \vspace*{-2.5pt} \item \ldots
\end{itemize}}};

\end{tikzpicture}

%% file: sections/ch3_data_driven_corner_cases.tex
\section{Method Layer Corner Cases}
\label{sec:data-droven-corner-cases}

In contrast to the corner cases of the \textit{sensor}, \textit{content}, and \textit{temporal layer}, there exists another corner case layer independent of the respective sensors, and the (actual) temporal and spatial content, which is not shown in Table~\ref{tab:cc_camera_lidar_radar_fusion}, as the respective corner cases are not sensor-, but method-specific. We term this corner case layer the \textit{method layer}, and refer to the respective corner cases as \textit{method level} corner cases. These corner cases are not necessarily perceptible by sensors or a human driver, making this layer more abstract.
   
Corner cases of the \textit{method layer} are caused by the applied method themselves. In the processing toolchain in Figure \ref{fig:ml_toolchain}, this corner case type appears predominantly in all phases involving machine learning methods (i.e., phases III and V). Depending on their topology, design, and deployment, methods can give rise to various corner cases. These might be corner cases due to a lack of knowledge because the ML model has never encountered a similar situation before. In literature, this is also referred to as epistemic uncertainty \cite{Kendall2017}, which can be approximate via Monte Carlo dropout \cite{Gal2016}, or deep ensembles \cite{Lakshminarayanan2017}.
As an example for a \textit{method level} corner case, consider a ML method for visual object detection. While we might aim to detect unknown objects by a high epistemic uncertainty \cite{Heidecker2021}, also typical ``normal'' objects can spark a high epistemic uncertainty \cite{Kendall2017,Mukhoti2018}, hence, leading to a \textit{method level} corner case. Epistemic uncertainty is not necessarily limited to ML methods and can appear in various kinds of mathematical models in general.
Adversarial samples are another example of \textit{method level} corner cases that are not perceptible by humans (e.g. \cite{Szegedy2014,Goodfellow2015}). Even small changes to the model's input (e.g., adding a particular noise pattern) can lead to drastic changes in the model's output.  
One challenge is to distinguish corner cases of the \textit{sensor}, \textit{content}, and \textit{temporal layer} from corner cases only immanent to the \textit{method layer}.

Depending on the method and model, an inductive bias, i.e. assumptions inherent to the ML method influencing its predictive capability, is introduced to the underlying perception task. This bias is also connected to \textit{method level} corner cases as it introduces certain assumptions about the problem at hand, hence, also manifesting the types of corner cases that can appear. 

There is a semantic overlap between \textit{method level} corner cases and the \textit{single-source} corner cases resulting from the above explications. While \textit{method level} corner cases are caused by uncertainty in the methodology, this definition does not exclude other corner case types. If we encounter a bear on the street, this can be considered an \textit{object level} corner case for both RADAR and LiDAR. However, at the same time, it is possible that the appearance of this bear also results in high epistemic uncertainty \cite{Mukhoti2018}. The same holds, e.g., if one encounters a \textit{domain level} corner case through a location change (see Table~\ref{tab:cc_camera_lidar_radar_fusion}), depending on the method, this can also lead to high epistemic uncertainty.

%% file: sections/ch7_methods.tex
\section{Corner Case Detection Techniques}\label{sec:methods_for_detection}
    \jb{In this section, we give a brief overview of existing methods for corner case detection. Most corner case detection methods in literature regard the camera sensor and aim to detect corner cases in images or image sequences; see the overviews given in \cite{Breitenstein2020,Breitenstein2021}, we consider methods concerning different sensors, layers of corner cases, and their engagement with the perception toolchain in Figure \ref{fig:ml_toolchain}.
    
    To detect corner cases on \textit{sensor layer}, Protopapadakis et al.\ \cite{Protopapadakis2017} map maritime RADAR data via stacked autoencoders to a feature space, where they employ density-based clustering to detect outliers. Chakravarthy et al.\ \cite{Chakravarthy2020} apply a CNN to raw RADAR data to extract features and classify them using an open-set classification method to detect unknowns in the waveform. 
    As an open-set classifier, they choose SV-Means \cite{Pavy2018}, which is built upon other existing open-set classifiers and formulated for the open-set problem of waveform classification. 
    Both approaches also extend to \textit{content} and \textit{temporal layer} corner cases. 
    
    Lis et al.\ \cite{Lis2019} regenerate synthetic images from segmentation masks to identify \textit{content layer} corner cases via the residual error between the real and regenerated image. There are also \textit{content layer} corner case detection methods on LiDAR data, such as Wong et al.\ \cite{Wong2019}, who introduce an open-set instance segmentation network on point clouds that identifies unknown points in an embedding space and groups them into unknown instances. Capellier et al.\ \cite{Capellier2019} propose a method to detect both known and unknown objects in LiDAR data.\\
    Bolte et al.\ \cite{Bolte2019b} identify \textit{temporal layer} corner cases for the camera sensor if the residual error between the real and a predicted image, weighted by the criticality of the location in the image, exceeds a threshold.
    
    For the camera sensor, methods from outlier and anomaly detection in more general computer vision can be applied to visual perception in automated driving. Of the three sensors considered in this work, corner case detection methods for RADAR data are the rarest. Adaptations of existing methods for LiDAR point clouds to RADAR data, e.g., the adaptation of PointNets to RADAR \cite{Danzer2019}, can also lead to new corner case detection methods in RADAR data.\\
    Most sensor fusion approaches aim to make the perception of automated vehicles more robust in general. Many focus on detecting obstacles, e.g., using infrared and ultrasonic sensors \cite{Farias2018}, or RADAR and LiDAR fusion with a Kalman-filter-type algorithm \cite{Hajri2018}. Yang et al.\ \cite{Yang2020} perform RADAR and LiDAR fusion via a novel neural network architecture leveraging early and late fusion for geometric and velocity data from the RADAR sensor, making object detection more robust.
    
}

%% file: sections/ch8_evaluation.tex
\newpage
\section{Evaluation}\label{sec:evaluation}
    \jb{In this section, we discuss existing datasets and methods for the evaluation of corner case detectors. Here, we especially elaborate on datasets and metrics to evaluate not only corner cases for visual perception, but those encompassing other sensors as well.}
    
    \subsection{Datasets}
        \jb{There exist a multitude of datasets for perception in highly automated driving. 
        
        \textbf{Camera:} Many of them provide camera images or sequences (cf. \cite{Breitenstein2020}), such as the \texttt{Cityscapes} dataset \cite{Cordts2016}, the \texttt{BDD100k} dataset \cite{Yu2018b}, the openDD dataset \cite{Breuer2020}, or the \texttt{Mapillary Vistas} dataset \cite{Neuhold2017}, to name a few. All of those are large-scale datasets for visual perception and provide adequate labels for the related tasks such as semantic and instance segmentation and object detection. Moreover, there exist a few datasets specifically for the task of unknown object detection from images, such as the \texttt{RoadAnomaly} \cite{Lis2019}, the \texttt{Lost\&Found} \cite{Pinggera2016} dataset, and the \texttt{Fishyscapes} \cite{Blum2019} dataset. Additionally, there exist datasets with a specific underlying task such as VRU detection (\texttt{Eurocity Persons}) \cite{Braun2019}.\\
        \textbf{LiDAR:} Next to those purely camera-based datasets, there exists by now also an increasing amount of multi-modal datasets. One of the earliest of such datasets is the \texttt{KITTI} \cite{Geiger2013} dataset, providing both camera and LiDAR data. The combination of LiDAR and camera data appears in several datasets, such as \texttt{Apolloscape} \cite{Huang2018}, \texttt{A2D2} \cite{Geyer2020}, \texttt{Waymo Open} \cite{Waymo2019}, \texttt{PandaSet} \cite{PandaSet}, and \texttt{KAIST} \cite{Jeong2019}. Many of them provide 3D bounding boxes as labels \cite{Waymo2019,Geyer2020,Geiger2013}, some additionally provide point cloud semantic segmentation labeling \cite{Geyer2020,Huang2018}. The \texttt{Canadian Adverse Driving Conditions} \cite{Pitropov2020} dataset specifically provides image and LiDAR data for wintry weather conditions.\\
        \textbf{RADAR:} RADAR data is rare in large-scale datasets for perception in automated driving. To our knowledge, only a few such datasets exist. \texttt{nuScenes} \cite{Caesar2019}, \texttt{Astyx HiRes2019} \cite{Meyer2019b}, \texttt{Oxford RobotCar} \cite{Maddern2017}, and \texttt{RADIATE} \cite{Sheeny2020} provide multi-modal data from RADAR, camera, and LiDAR, and as far as we know, all of these datasets provide bounding box labels for their RADAR data, except the \texttt{Oxford RobotCar} where no ground truth labels are available.}
        
    \subsection{Evaluation Methods}
        \jb{A critical aspect of corner case detection is how to evaluate such methods. In contrast to related fields in automotive perception, large-scale benchmarks are missing, and no consistent evaluation rules exist. Here, we aim to provide a brief overview of existing and applied methods to contribute to harmonizing this field of research.\\
        For corner case detection on camera images, often, the area under the receiver operator characteristic (AUC) is used to determine separability between a normal and anomalous corner case class (e.g., \cite{Chan2020}). Moreover, the area under the precision-recall curve (AUPRC) determines the separability of one class, e.g., the corner case class \cite{Chan2020}.
        One of few existing online benchmarks related to corner case detection on automotive camera images, the \texttt{Fishyscapes} online benchmark \cite{Blum2019}, reports results in the average precision (AP) and the false positive rate at 95\% true positive rate (FPR$_\text{95}$).\\
        For corner case detection on LiDAR point clouds, one often employs the task performance metrics average precision (AP) and panoptic quality (PQ) \cite{Wong2019}. Wong et al.\ \cite{Wong2019} additionally propose a variation of PQ, the unknown quality (UQ), which emphasizes recall quality over recognition quality.\\
        Next to the detection of corner cases, it is also important not to lose performance on the original perception task \cite{Chan2020}. Thus, one should report, e.g., accuracy, IoU (semantic segmentation), AP (object detection), especially when, e.g., including corner case detection and perception in a multi-task learning framework, where a learned corner case detection influences perception and vice versa.
        
        Finally, there exist corner case metrics, which follow specifically proposed corner case definitions. As an example, Bolte et al.\ \cite{Bolte2019b} propose a prediction-based corner case metric signifying that they obtain a corner case score from the error of an image prediction restricted to relevant objects in relevant locations.}
        

%% file: sections/ch9_conclusion.tex
\section{Conclusions and Future Work} \label{sec:conclusion}
In this work, we introduced a categorization of corner cases for perception in automated driving concerning the camera, LiDAR, and RADAR sensor modalities, highlighting specific definitions and respective examples for the sensors.  Additionally, we described a perception toolchain for automated driving, where the interfaces of corner case detection were highlighted in the pipeline. To comply with the multitude of sensors in automated driving, we introduced a \textit{sensor layer} for corner cases, where we distinguish the hardware and the \textit{physical level}. Moreover, \textit{method layer} corner cases were defined, which differ from the other corner case levels by being independent of the sensor modality, but instead are method-specific.

\addtolength{\textheight}{-0.0cm}   

%% file: sections/acknowledgment.tex
\section*{Acknowledgment}
    This work results from the project KI Data Tooling (19A20001O) funded by German Federal Ministry for Economic Affairs and Energy (BMWI) and the DeCoInt$^2$-project financed by the German Research Foundation (DFG) within the priority program SPP 1835: “Kooperativ interagierende Automobile”, grant number SI 674/11-2.